# Benchmarking Open-Source Large Language Models on Healthcare Text Classification Tasks

**Yuting Guo, MS[1], Abeed Sarker, PhD[1]**
**[1]Department of Biomedical Informatics, School of Medicine, Emory University, Atlanta, GA**

**Abstract**
*The application of large language models (LLMs) to healthcare information extraction has emerged as a promising approach for automating text classification tasks across diverse data sources. This study evaluates the classification performance of five open-source LLMs: GEMMA-3-27B-IT, LLAMA3-70B, LLAMA4-109B, DEEPSEEK-R1-DISTILL-LLAMA-70B, and DEEPSEEK-V3-0324-UD-Q2_K_XL, across six healthcare-related classification tasks involving both social media data (breast cancer, changes in medication regimen, adverse pregnancy outcomes, potential COVID-19 cases) and clinical data (stigma labeling, medication change discussion). We report precision, recall, and $F_1$ scores with 95% confidence intervals for all model-task combinations. Our findings reveal significant performance variability between LLMs, with DeepSeekV3 emerging as the strongest overall performer, achieving the highest $F_1$ scores in four tasks. Notably, models generally performed better on social media tasks compared to clinical data tasks, suggesting potential domain-specific challenges. GEMMA-3-27B-IT demonstrated exceptionally high recall despite its smaller parameter count, while LLAMA4-109B showed surprisingly underwhelming performance compared to its predecessor LLAMA3-70B, indicating that larger parameter counts do not guarantee improved classification results. We observed distinct precision-recall trade-offs across models, with some favoring sensitivity over specificity and vice versa. These findings highlight the importance of task-specific model selection for healthcare applications, considering the particular data domain and precision-recall requirements rather than model size alone. As healthcare increasingly integrates AI-driven text classification tools, this comprehensive benchmarking provides valuable guidance for model selection and implementation while underscoring the need for continued evaluation and domain adaptation of LLMs in healthcare contexts.*[1]

***Keywords:*** *DeepSeek, Llama, text classification, large language models, natural language processing,*

## Introduction

Large language models (LLMs) have shown significant promise in various healthcare applications, including medical text classification.[1,2] As these models continue to evolve, it becomes increasingly important to benchmark their performance on healthcare-specific tasks to understand their capabilities and limitations[2]. The biomedical domain presents unique challenges for natural language processing due to specialized terminology, contextual nuances, and the high importance of accuracy. Understanding how different models perform on these tasks can guide practitioners in selecting appropriate models for specific healthcare applications.

Although closed-access large language models (LLMs) such as the GPT series, Grok, and Gemini have garnered substantial research interest, their closed-source nature and lack of HIPAA compliance often render them unsuitable for use with medical texts. Consequently, it is essential to conduct LLM research using open-source models when working with medical data to ensure compliance and reproducibility. This paper benchmarks five relatively new open-source LLMs: GEMMA-3-27B-IT,[3] LLAMA3-70B,[4] LLAMA4-109B,[5] DEEPSEEK-R1-DISTILL-LLAMA-70B,[6] and DEEPSEEK-V3-0324-UD-Q2_K_XL.[7] The benchmark includes six classification tasks with social media health content and clinical notes, providing insights into how these models perform across different healthcare text domains.

## Methods

We chose six biomedical text classification tasks for benchmarking—four involving social media data (X/Twitter posts) focusing on detection of self-reported breast cancer,[8] changes in medication regimen,[9] adverse pregnancy outcomes,[10] and potential cases of COVID-19,[11] and two tasks involving data from clinical notes in electronic health records (EHRs) focusing on stigmatizing language detection,[12] and the presence of a medication change discussion.[13] All of the six tasks involve binary text classification. The data statistics for all tasks are shown in Table 1.

We evaluated five LLMs as follows:

---

[1] We are actively benchmarking new LLMs and will release the lates results at https://sarkerlab.org/benchmark.

GEMMA-3-27B-IT

GEMMA-3-27B-IT[2] is a 27-billion-parameter instruction-tuned model developed to support high-quality performance in interactive and task-specific natural language understanding. It is designed to follow user prompts effectively across a range of applications, including question answering, summarization, and dialogue generation. Its instruction tuning enables improved alignment with human intent, making it suitable for real-world deployment.

LLaMA3-70B

LLaMA3-70B[3] is part of Meta's third-generation LLaMA series, comprising 70 billion parameters and built to offer enhanced language understanding, reasoning, and generalization capabilities. It demonstrates strong performance on a wide range of benchmarks, rivaling or exceeding many closed-source models. LLaMA3-70B is open-weight and has become a popular choice for research and development in safety-aligned and transparent AI.

LLaMA4-109B

LLaMA4-109B[4] is an upcoming or research-based extension of the LLaMA model family, scaling to 109 billion parameters for greater expressivity and contextual understanding. It is designed to address complex reasoning and long-context tasks while maintaining efficient training dynamics. As with earlier versions, it emphasizes openness and reproducibility for academic and industrial use.

DeepSeek-R1-Distill-LLaMA-70B

DeepSeek-R1-Distill-LLaMA-70B[5] is a distilled version of a 70-billion-parameter LLaMA-based model, aimed at achieving a balance between performance and efficiency. Through distillation, the model retains much of the predictive power of its larger counterpart while enabling faster inference and lower resource consumption. This makes it particularly attractive for deployment in real-time or resource-constrained environments.

DeepSeek-V3-0324-UD-Q2_K_XL

DeepSeek-V3-0324-UD-Q2_K_XL[6] is a quantized and optimized variant with 685 billion parameters in the DeepSeek series, tailored for ultra-dense understanding (UD) tasks. The model is designed for efficient inference without sacrificing performance, making it well-suited for downstream applications such as summarization, classification, and retrieval-based tasks. The Q2_K format indicates a quantization strategy to reduce memory footprint while maintaining model accuracy.

For each LLM, the model received a plain text prompt with detailed instructions and output formatting guidelines to facilitate post-processing. Each dataset was divided 80/20 into training and test sets—the former for prompt optimization and the latter for performance evaluation. We focused on minimizing invalid predictions and generalizing the prompts over the six classification tasks rather than exhaustive prompt exploration, concluding optimization when invalid answers numbered fewer than 10. All evaluation experiments on the 20% test data were conducted in zero-shot settings. 95% confidence intervals were computed using bootstrap resampling.[11] Table 2 displays the task-specific prompts.

**Table 1.** Data statistics for the six classification tasks.

| Task | Positive | Negative | Total |
|---|---|---|---|
| Breast cancer | 1283 (26%) | 3736 (74%) | 5019 |
| Changes in medication regimen | 656 (9%) | 6814 (91%) | 7470 |
| Adverse pregnancy outcomes | 2922 (45%) | 3565 (55%) | 6487 |
| Potential cases of COVID-19 | 1148 (16%) | 6033 (84%) | 7181 |
| Stigma labeling | 439 (44%) | 560 (56%) | 999 |

[2] Downloaded from https://huggingface.co/google/gemma-3-27b-it on April 11, 2025.
[3] Downloaded from https://huggingface.co/meta-llama/Meta-Llama-3-70B-Instruct on March 25, 2025.
[4] Downloaded from https://huggingface.co/meta-llama/Llama-4-Scout-17B-16E-Instruct on April 9, 2025.
[5] Downloaded from https://huggingface.co/deepseek-ai/DeepSeek-R1-Distill-Llama-70B on March 28 , 2025.
[6] Downloaded from https://huggingface.co/unsloth/DeepSeek-V3-0324-GGUF. Accessed on April 8, 2025.

| | | | |
|---|---|---|---|
| Medication change discussion | 1327 (20%) | 5232 (80%) | 6559 |

**Table 2.** Prompt templates for classification tasks. For social media tasks, the template replaced '{{task}}' with the specific task (breast cancer, medication regimen changes, adverse pregnancy outcomes, or potential COVID-19 cases) and '{{text}}' with the post content. For clinical note tasks, '{{text}}' was replaced with the note content.

| **Task** | **Prompt Template** |
|---|---|
| Social media tasks | Read the following tweet. Answer if the user is self-reporting {{task}} and explain why.<br>Tweet:{{text}}<br>Your answer should state "the tweet is self-reporting {{task}}" or "the tweet is not self-reporting potential". |
| Stigma labeling | Read the following clinical note.<br>Clinical note:{{text}}<br>Answer if this note involves language about the patient which could result in the stigmatization or negative labeling of a patient, which could lead to further status loss/discrimination in the context of the patient-provider relationship. Your answer should state "the note involves ..." or "the note does not involve ...". |
| Medication change discussion | Read the following clinical note.<br>Clinical note:{{text}}<br>Answer if the presence of a medication change is discussed. Your answer should state "the presence of a medication change is discussed" or "the presence of a medication change is not discussed". |

**Results**

**Table 3.** The classification performance of five LLMs over six tasks. The precision, recall, and $F_1$ score with 95% confidence intervals are reported.
Abbreviations: Gemma3=GEMMA-3-27B-IT, Llama3=LLAMA3-70B, Llama4=LLAMA4-109B, DeepSeekR1=DEEPSEEK-R1-DISTILL-LLAMA-70B, DeepSeekV3=DEEPSEEK-V3-0324-UD-Q2_K_XL.

| Task | Precision | | | | |
|---|---|---|---|---|---|
| | Gemma3 | Llama3 | Llama4 | DeepSeekR1 | DeepSeekV3 |
| Breast cancer | 0.71<br>(0.66-0.75) | 0.84<br>(0.80-0.88) | 0.76<br>(0.67-0.84) | **0.87**<br>(0.83-0.91) | **0.87**<br>(0.84-0.91) |
| Changes in medication regimen | 0.13<br>(0.11-0.15) | 0.27<br>(0.22-0.33) | 0.19<br>(0.14-0.23) | 0.30<br>(0.24-0.35) | **0.38**<br>(0.32-0.44) |
| Adverse pregnancy outcomes | 0.55<br>(0.52-0.58) | 0.72<br>(0.68-0.75) | 0.67<br>(0.62-0.73) | **0.73**<br>(0.69-0.76) | 0.71<br>(0.68-0.75) |
| Potential cases of COVID-19 | 0.47<br>(0.41-0.53) | 0.66<br>(0.59-0.74) | 0.30<br>(0.24-0.37) | 0.64<br>(0.55-0.72) | **0.73**<br>(0.65-0.80) |
| Stigma labeling | 0.64<br>(0.52-0.75) | 0.52<br>(0.44-0.60) | 0.51<br>(0.42-0.60) | **0.67**<br>(0.57-0.75) | 0.60<br>(0.49-0.71) |
| Medication change discussion | 0.33<br>(0.29-0.36) | 0.40<br>(0.36-0.44) | 0.37<br>(0.32-0.41) | 0.26<br>(0.23-0.29) | **0.46**<br>(0.42-0.51) |

| Task | Recall | | | | |
|---|---|---|---|---|---|
| | Gemma3 | Llama3 | Llama4 | DeepSeekR1 | DeepSeekV3 |
| Breast cancer | **0.94**<br>(0.91-0.97) | 0.92<br>(0.88-0.94) | 0.24<br>(0.19-0.29) | 0.92<br>(0.88-0.95) | 0.92<br>(0.89-0.95) |
| Changes in medication regimen | **0.93**<br>(0.88-0.97) | 0.49<br>(0.41-0.57) | 0.38<br>(0.30-0.46) | 0.60<br>(0.51-0.68) | 0.59<br>(0.51-0.67) |
| Adverse pregnancy outcomes | **0.92**<br>(0.90-0.94) | 0.82<br>(0.79-0.85) | 0.34<br>(0.31-0.39) | 0.83<br>(0.80-0.86) | 0.86<br>(0.84-0.89) |
| Potential cases of COVID-19 | **0.60** (0.54-0.66) | 0.40 (0.33-0.46) | 0.23 (0.17-0.28) | 0.33 (0.27-0.39) | 0.39 (0.33-0.45) |
| Stigma labeling | 0.48 (0.38-0.58) | **0.91** (0.84-0.96) | 0.69 (0.60-0.79) | 0.70 (0.61-0.80) | 0.61 (0.51-0.71) |
| Medication change discussion | **0.97** (0.95-0.99) | 0.84 (0.79-0.88) | 0.61 (0.55-0.67) | 0.82 (0.77-0.87) | 0.77 (0.73-0.82) |

| Task | $F_1$ | | | | |
|---|---|---|---|---|---|
| | Gemma3 | Llama3 | Llama4 | DeepSeekR1 | DeepSeekV3 |
| Breast cancer | 0.81 (0.77-0.84) | 0.88 (0.85-0.90) | 0.36 (0.30-0.42) | 0.89 (0.86-0.92) | **0.90** (0.87-0.92) |
| Changes in medication regimen | 0.23 (0.19-0.26) | 0.35 (0.28-0.41) | 0.25 (0.19-0.30) | 0.40 (0.34-0.46) | **0.46** (0.40-0.53) |
| Adverse pregnancy outcomes | 0.69 (0.66-0.72) | 0.77 (0.74-0.79) | 0.46 (0.42-0.50) | 0.77 (0.75-0.80) | **0.78** (0.76-0.80) |
| Potential cases of COVID-19 | **0.53** (0.47-0.58) | 0.50 (0.43-0.55) | 0.26 (0.20-0.32) | 0.44 (0.37-0.50) | 0.51 (0.45-0.57) |
| Stigma labeling | 0.55 (0.45-0.64) | 0.66 (0.59-0.73) | 0.59 (0.51-0.67) | **0.69** (0.60-0.76) | 0.61 (0.52-0.69) |
| Medication change discussion | 0.49 (0.45-0.52) | 0.54 (0.50-0.58) | 0.46 (0.41-0.50) | 0.39 (0.36-0.43) | **0.58** (0.53-0.62) |

Table 3 presents the precision, recall, and $F_1$ scores with 95% confidence intervals for five LLMs across six classification tasks. The results for each task are described as follows:

**Breast cancer:** All models achieved strong performance, with DeepSeekV3 yielding the highest $F_1$ score (0.90), closely followed by DeepSeekR1 (0.89) and Llama3 (0.88). Although Llama4 showed a high precision (0.76), its recall was notably low (0.24), resulting in a substantially reduced $F_1$ score (0.36).

**Changes in medication regimen:** Performance was comparatively lower across all models. DeepSeekV3 achieved the best $F_1$ score of 0.46, showing a balanced precision (0.38) and recall (0.59). Other models underperformed, particularly Gemma3 and Llama4, which had $F_1$ scores of 0.23 and 0.25, respectively, due to poor precision.

**Adverse pregnancy outcomes:** DeepSeekV3 led with an $F_1$ score of 0.78, followed closely by Llama3 and DeepSeekR1 (both at 0.77). Llama4 performed substantially worse ($F_1$ = 0.46), again due to low recall (0.34).

**Potential cases of COVID-19:** This task revealed moderate performance. DeepSeekV3 obtained the highest $F_1$ score (0.51), while Llama4 struggled significantly ($F_1$ = 0.26), again attributable to low recall (0.23). Gemma3 and DeepSeekR1 showed intermediate results with $F_1$ scores of 0.53 and 0.44, respectively.

**Stigma labeling:** DeepSeekR1 achieved the best $F_1$ score (0.69), followed by Llama3 (0.66). Gemma3 and Llama4 showed lower performance with $F_1$ scores of 0.55 and 0.59, respectively, while DeepSeekV3 showed moderate performance ($F_1$ = 0.61).

**Medication change discussion:** DeepSeekV3 again outperformed others with an F1 score of 0.58. While Llama3 and Gemma3 achieved $F_1$ scores of 0.54 and 0.49, Llama4 and DeepSeekR1 were less effective, with scores of 0.46 and 0.39, respectively.

Overall, DeepSeekV3 demonstrated the most consistently strong performance across tasks, particularly in handling clinical nuances in adverse pregnancy outcomes and medication change discussions. Llama3 also showed robust performance in multiple tasks, while Llama4, despite strong precision in some cases, was hindered by significantly lower recall.

**Discussion**

The findings highlight several key trends in model behavior, as well as practical implications for LLM deployment in medical and health-related natural language processing. Across the social media tasks (i.e., breast cancer, changes in medication regimen, adverse pregnancy outcomes, and potential COVID-19 cases), the best-performing model was consistently DeepSeekV3, which demonstrated strong generalization capabilities despite the informal and noisy nature of user-generated content. These results underscore the importance of model scale when applying LLMs to non-standardized language data typical of social platforms. Notably, Llama4, despite its larger parameter size, underperformed in nearly all tasks. Its consistently low recall, especially in the breast cancer and adverse pregnancy outcome tasks, suggests that it may be over-conservative in predictions or insufficiently tuned for recall-sensitive domains. This highlights that larger model size alone does not guarantee better performance, particularly when models are not specifically adapted to the target domain or task.

In contrast, the clinical text tasks (i.e., stigma labeling and medication change discussion) presented unique challenges related to context sensitivity and medical domain complexity. DeepSeekR1 and DeepSeekV3 outperformed all other models achieving the highest $F_1$ scores (0.61 for stigma labeling and 0.58 for medication change discussion, respectively). Their superior performance may be attributed to the efficient representation learning that retains critical semantic details even under limited context. Additionally, Llama3 maintained strong performance, particularly in stigma labeling ($F_1$ = 0.66), indicating its capacity to handle nuanced clinical concepts when properly instructed.

These findings have several important implications for the application of LLMs in healthcare contexts. First, the choice of LLM should be task-specific rather than based solely on parameter count or general reputation. For instance, Gemma3 may be preferable for high-recall requirements despite its smaller size. Second, the performance gap between social media and clinical data classification suggests that domain-specific fine-tuning might be necessary to optimize LLM performance for clinical applications. Third, for healthcare applications where false negatives could have serious consequences (e.g., missing adverse pregnancy outcomes), models with higher recall like Gemma3 might be preferred despite lower precision. Conversely, applications where false positives are costly might benefit from higher-precision models. Finally, the varying strengths of different models across tasks suggest potential benefits from ensemble approaches that leverage multiple LLMs.

Future work should explore multi-domain training strategies, task-specific prompting, and further evaluation on real-world clinical tasks. Additionally, greater emphasis should be placed on evaluating recall, especially in applications where false negatives could have critical implications, such as adverse event detection or patient safety monitoring.

## Limitations

Several limitations should be acknowledged. First, the confidence intervals for certain metrics are quite wide, indicating potential instability in model performance. Second, the binary classification approach may oversimplify the complexity of healthcare information. Future work should explore more nuanced multi-class classification or sequence labeling approaches. Additionally, investigating the impact of prompt engineering and few-shot learning on classification performance could yield valuable insights. Exploring model interpretability to understand the linguistic features that drive classification decisions would also enhance trust in these systems for healthcare applications.

## Conclusions

This study evaluated five open-source LLMs across six healthcare-related classification tasks spanning social media and clinical data domains. DeepSeekV3 demonstrated the strongest overall performance, achieving the highest $F_1$ scores in four tasks, while Gemma3 showed exceptional recall despite its smaller size, and DeepSeekR1 excelled in clinical stigma detection. Notably, larger parameter counts did not guarantee better performance, as evidenced by Llama4's underperformance compared to Llama3. Models generally performed better on social media tasks than clinical data tasks, suggesting domain-specific challenges. These findings underscore that optimal model selection for healthcare applications should be task-specific, considering the particular data domain and precision-recall requirements rather than model size alone. As healthcare increasingly incorporates AI-driven text classification, continued benchmarking and domain adaptation will be essential to ensure effective and responsible implementation of these powerful but still-evolving technologies.